\documentclass[runningheads]{llncs}
\usepackage{makeidx}  
\usepackage{etoolbox} 
\makeatletter
\patchcmd{\ps@headings}{\rlap{\thepage}}{}{}{}
\patchcmd{\ps@headings}{\llap{\thepage}}{}{}{}
\makeatother
\usepackage{orcidlink}
\usepackage{hyperref} 
\usepackage{multirow}
\usepackage{caption}
\usepackage{subcaption}
\usepackage{amsmath} 
\usepackage{dsfont}  
\usepackage{longtable}
\usepackage{todonotes}
\usepackage{geometry}
\usepackage{makecell} 
\usepackage{multirow}
\usepackage{booktabs} 

\usepackage{todonotes}

\usepackage{comment}

\hypersetup{colorlinks=false,%
            urlbordercolor=blue,
            pdfborderstyle={/S/U/W 0.5}}

\begin{document}

\mainmatter 
\title{Supervised and self-supervised land-cover \\ segmentation \& classification of the Biesbosch wetlands}
\titlerunning{Land-cover classification \& segmentation of the Biesbosch wetlands}
\author{
Eva {Gmelich Meijling}
\inst{1,3} 
\and 
Roberto Del Prete\orcidID{\href{https://orcid.org/0000-0003-0810-4050}{\includegraphics[scale=0.05]{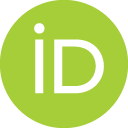}}}\inst{1,2}
\and 
Arnoud Visser \orcidID{\href{https://orcid.org/0000-0002-7525-7017}{\includegraphics[scale=0.05]{ORCIDiD_icon128x128.png}}}\inst{3}
}

\authorrunning{
{E.} {Gmelich Meijling} 
et al.}

\institute{
$\Phi$-lab ESRIN, European Space Agency, Via Galileo Galilei 1, 00044 Frascati, Italy \and
Dept. of Aerospace Engineering, University of Naples Federico II, 80125 Napoli, Italy \and
University of Amsterdam, Science Park 900, 1098 XH Amsterdam, The Netherlands}

\maketitle



\begin{abstract}



Accurate wetland land-cover classification is essential for environmental monitoring, biodiversity assessment, and sustainable ecosystem management. However, the scarcity of annotated data, especially for high-resolution satellite imagery, poses a significant challenge for supervised learning approaches. To tackle this issue, this study presents a methodology for wetland land-cover segmentation and classification that adopts both supervised and self-supervised learning (SSL). We train a U-Net model from scratch on Sentinel-2 imagery across six wetland regions in the Netherlands, achieving a baseline model accuracy of 85.26\%.

Addressing the limited availability of labeled data, the results show that SSL pretraining with an autoencoder can improve accuracy, especially for the high-resolution imagery where it is more difficult to obtain labeled data, reaching an accuracy of 88.23\%.

Furthermore, we introduce a framework to scale manually annotated high-resolution labels to medium-resolution inputs. While the quantitative performance between resolutions is comparable, high-resolution imagery provides significantly sharper segmentation boundaries and finer spatial detail.

As part of this work, we also contribute a curated Sentinel-2 dataset with Dynamic World labels, tailored for wetland classification tasks and made publicly available\footnote{\scriptsize  \href{https://doi.org/10.5281/zenodo.15125549}{DOI: 10.5281/zenodo.15125549}}.

\end{abstract}

\section{Introduction}

Wetlands face growing threats from human development and climatic impacts on their ecosystems \cite{salimi2021impact}. Monitoring these environments is crucial for understanding changes and making informed decisions to mitigate or reduce these threats. If not properly observed, changes in wetlands can lead to disruptions in surrounding areas, such as increased flooding risks, loss of biodiversity, and degradation of water quality. Jafarzadeh et al. \cite{Jafarzadeh2022RemoteResearch} reviewed 334 studies on wetland monitoring using remote sensing (RS) over the past three decades. More than half of the studies focused on classifying the region as wetland zone. 
A subset of these studies explored  the internal changes of the wetland zone, including vegetation mapping, and the delineation of wetland extent. Similarly, this study investigates internal wetland dynamics, focusing specifically on vegetation mapping within a wetland area in the Netherlands.

\subsection{Context}

\begin{figure}[ht]
    \centering
    \includegraphics[width=0.5\linewidth]{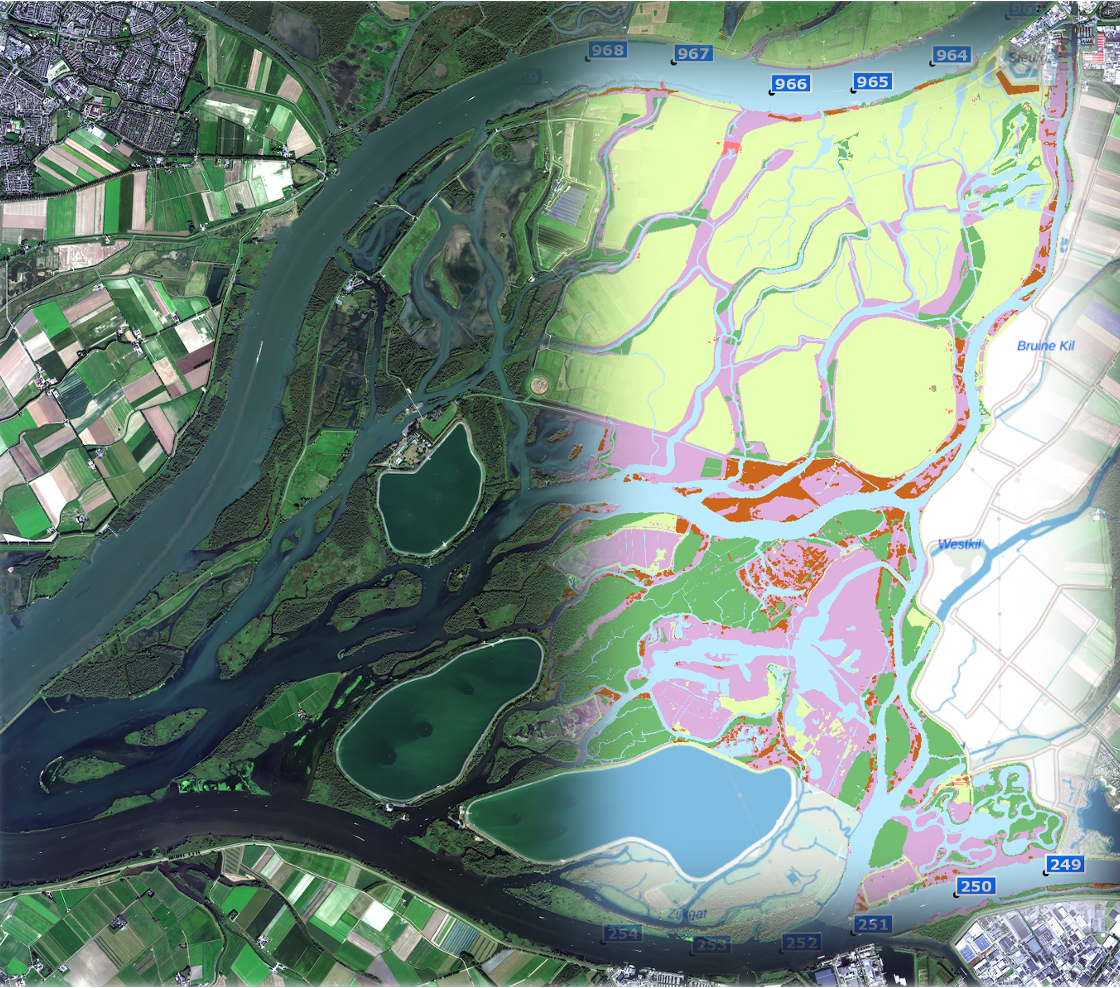}
    \caption{\textit{\small Sentinel-2 satellite image of the Biesbosch region with a conceptual land-cover classification overlay (artist impression by Rijkswaterstaat).}}
    \label{fig:Biesbosch}
\end{figure}

The focus of this study is the Biesbosch region, shown in Figure~\ref{fig:Biesbosch}. It is one of Europe's few remaining freshwater tidal areas, shaped by the St. Elisabeth's Flood in 1421. 
The area underwent its most dramatic hydrological change with the Delta Works Project. After the completion of the Haringvlietdam in 1970 the tidal differences reduced from variances as large as 2 meters to variances between 20-80 centimeters. Despite these changes, the Biesbosch remains one of the largest and most valuable natural areas in the Netherlands, covering approximately 9,700 hectares \cite{zhu2020historic}.

The distribution and density of the area's vegetation strongly influence the water flow patterns, making accurate classification of vegetation essential to effectively monitor and manage the ecosystem. Monitoring the area helps identify locations where water builds up due to excessive or overly rough vegetation. As a large and relatively inaccessible areas that frequently undergo rapid and unpredictable environmental changes \cite{Gallant2015TheWetlands}, RS with satellite images is an efficient method to observe and track the dynamics of the landscape. Based on the remote observation, action can be taken such as mowing by humans or deploying natural grazers.

\subsection{Related work}
The ability of Artificial Intelligence (AI) methods to detect patterns and trends in large datasets has seen significant advancements. The development of deep learning (DL) methods, such as multilayered neural networks, vision transformers (ViTs), generative adversarial networks (GANs), and large-scale visual segmentation models, has greatly enhanced the capacity of AI in RS \cite{Janga2023ASciences}. These advancements allow for more accurate and actionable insights that were previously difficult to achieve with traditional RS methods. 

Although DL methods have shown 16–21\% higher accuracy than traditional approaches in land-cover and wetland classification by capturing complex patterns in RS data \cite{Mainali2023ConvolutionalModel}, the data-hungry paradigm of DL is not always the preferred choice \cite{Jafarzadeh2022RemoteResearch} due: a) interpretation challenges, b) large data requirements, and c) high computational costs. The review \cite{Jafarzadeh2022RemoteResearch} of 344 studies (1990–2022), mostly published in the \textit{Remote Sensing} journal, found that ensemble learning, particularly Random Forest (RF), remains the most widely used ML approach for wetland research due to its robustness with multi-source RS data.

With the introduction of foundation models (FM) in 2021, a new paradigm shift took place in AI-driven RS \cite{lu2025vision}. Unlike task-specific models, which are optimized for specific applications, foundation models, such as CROMA \cite{fuller2024croma}, are pretrained on a diversity of datasets, allowing them to generalize across multiple RS tasks with minimal fine-tuning. As highlighted by Lu et al. \cite{lu2025vision}, these models excel in transfer learning, few-shot adaptation, and multi-modal integration, making them highly effective for diverse RS tasks. However, their limitations include domain gaps between natural image datasets \cite{rs16050797}, which are often pre-trained and require task-specific fine-tuning for optimal performance. Furthermore, similar to other DL methods, FMs pose challenges in terms of interpretability \cite{10834497} and computational demands \cite{lu2025vision}.

To support the detailed classification of vegetation types, this study adopts a semantic segmentation framework, which enables both accurate categorization and preservation of spatial patterns within heterogeneous landscapes. The comparative analysis focuses on supervised and self-supervised learning strategies, evaluated within the ecologically diverse and hydrologically dynamic Biesbosch wetland region.

\section{Autoencoder \& U-Net architecture}
\label{sec:unet}
For semantic segmentation we chose for a U-Net architecture \cite{ronneberger2015u}. A U-Net has a symmetric encoder-decoder structure with skip connections to retain fine-grained spatial details. The encoder extracts compact feature representations by downsampling the input, while its decoder reconstructs detailed segmentation maps through upsampling. It's skip connections bridge corresponding encoder and decoder layers, enabling the model to retain both high-level abstract features and fine-grained spatial details. This design makes U-Net highly effective for tasks requiring precise segmentation, including delineating features in satellite images, such as vegetation boundaries \cite{Butko2024VegetationImagery,Dahiya2024SatelliteU-Net}. 

\begin{figure}[hbt]
    \centering
    \includegraphics[width=0.7\linewidth]{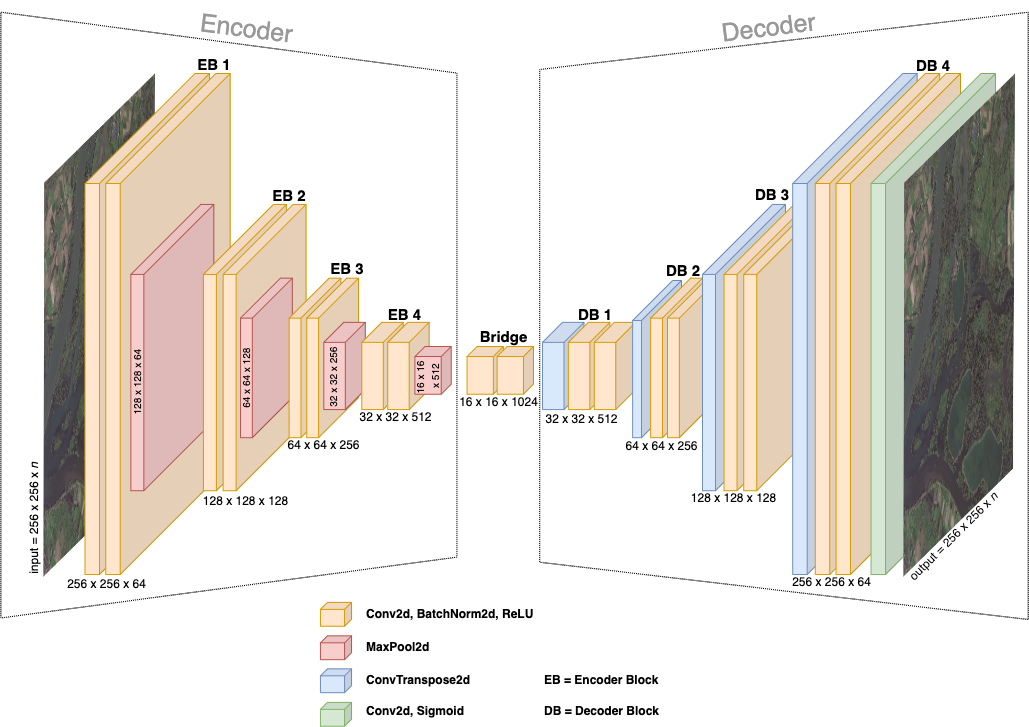}
    \caption{\textit{\small Schematic representation of the autoencoder architecture for 256×256 pixel input. The encoder reduces the spatial dimensions by a factor of 2 at each block, compressing the input from 256×256 pixels to 16×16 pixels in the bridge. At the same time, the number of channels increases by a factor of 2, from 64 in the first encoder block, to 512 in the bridge. The decoder restores the spatial dimensions 
    back to 256×256 pixels.}}
    \label{fig:AutoencoderArchitecture}
\end{figure}

U-Net architectures are frequently integrated with high-capacity pretrained encoders, such as ResNet or EfficientNet, to enhance feature extraction and improve segmentation performance \cite{he2016deep,pmlr-v97-tan19a}. Nonetheless, these standard backbones are typically trained on natural image datasets, which differ significantly from RS data. In this study, the models are trained from scratch, and we investigate whether the weights of an autoencoder trained on RS data can capture its unique spectral and spatial characteristics. The autoencoder used here is a fully convolutional architecture designed for image reconstruction. It consists of two main components: an encoder and a decoder, both composed of convolutional blocks for feature extraction and reconstruction. Figure~\ref{fig:AutoencoderArchitecture} provides a schematic overview of the autoencoder architecture.

To accommodate different image resolutions, the input data was divided into patches of varying sizes. For medium-resolution imagery, patches of \(256 \times 256\) pixels were used with a batch size of 8. For very-high-resolution imagery, larger patches of \(1024 \times 1024\) pixels were used, requiring a batch size of 4 due to higher memory demands.

To optimize autoencoder training performance, various learning rates and dropout probabilities were evaluated. A fixed learning rate of 0.001 and a dropout probability of 15\% yielded the best results among the tested values (0\%, 15\%, and 25\%) \cite{Eva2025thesis}.

\begin{figure}[hbt]
    \centering
    \includegraphics[width=0.7\linewidth]{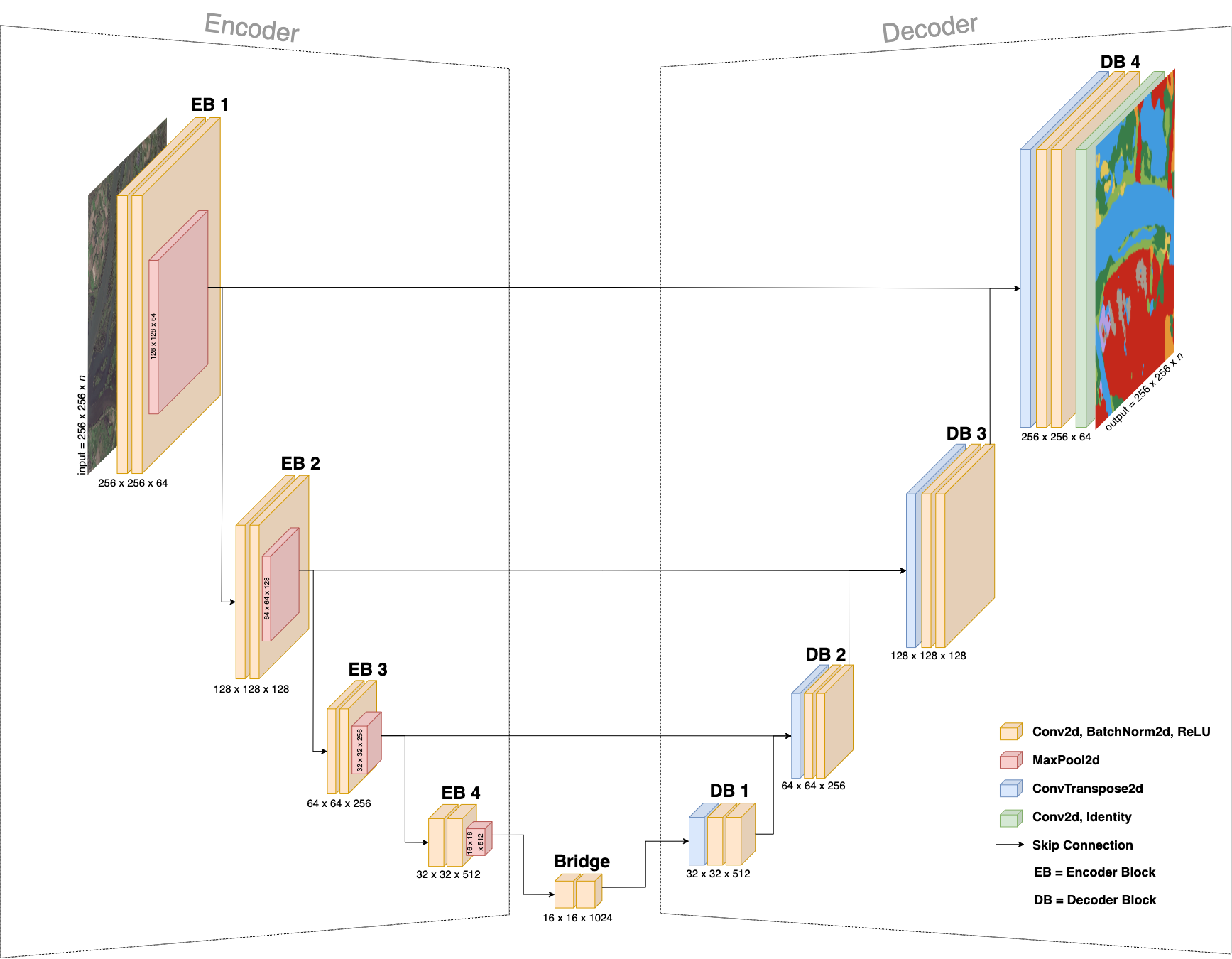}
    \caption{\textit{\small Schematic representation of the U-Net architecture for 256×256 pixel input. The encoder reduces spatial dimensions to 1024 channels of 16×16 pixels in the bridge. Skip connections link corresponding encoder and decoder layers, helping to retain spatial information. The decoder reconstructs the segmentation map, restoring the spatial dimensions back to 256×256.}}
    \label{fig:UnetArchitecture}
\end{figure}

The U-Net architecture shares the same encoder design as the autoencoder, which allows the weights to be transferred between the two models. It differs from the autoencoder by incorporating skip connections, which help retain fine-grained spatial details necessary for accurate segmentation. Figure~\ref{fig:UnetArchitecture} provides a schematic overview of the U-Net architecture. For U-Net training, cosine annealing yielded the best results for the learning rate \cite{Eva2025thesis}, combined with a dropout rate of 15\%.

\section{Methodology}

As stated above, the study uses a U-Net model to perform semantic segmentation for land-cover classification from satellite imagery. More in depth, to improve its performance, we investigate two key factors:

\begin{itemize}
    \item \textbf{Dependency on labeled data:} The performance of semantic segmentation models heavily depends on the labeled training data. Oftentimes, the annotated data is scarce or misaligned with the data intended to be used during inference, such as the desired classification scheme, resolution, and geographical coverage. This experiment addresses this limitation by exploring the use of widely available unlabeled RS data through pretraining.
    \item \textbf{Dependency on resolution:} Higher-resolution images provide more detailed spatial information, which could potentially improve model accuracy and reduce the dependency on labeled data. This experiment investigates how different image resolutions (medium resolution vs. very high resolution) affect the performance of the U-Net model under both from-scratch and pretrained conditions.
\end{itemize}

To enable and compare model training and evaluation across different resolutions and levels of label availability, this research used both medium-resolution and very-high-resolution optical satellite data obtained from open-source platforms, including Google Earth Engine\footnote{\scriptsize \url{https://earthengine.google.com}} and Satellietdataportaal\footnote{\scriptsize \url{https://viewer.satellietdataportaal.nl/}}. 
The Sentinel-2 dataset constructed for this study, including selected spectral bands, preprocessed tiles, and corresponding Dynamic World labels, has been made publicly available  [\href{https://doi.org/10.5281/zenodo.15125549}{DOI: 10.5281/zenodo.15125549}]. This contribution supports reproducibility and benchmarking for remote sensing in wetland environments. Due to licensing restrictions, the very-high-resolution (VHR) satellite imagery obtained from Satellietdataportaal cannot be shared.

\vspace*{0.25cm}
\textbf{Sentinel-2 imagery}\footnote{\scriptsize \url{https://www.esa.int/Applications/Observing_the_Earth/Copernicus/Sentinel-2}} was obtained via the Google Earth Engine API, where we selected Harmonized Level-2A Sentinel-2 data to ensure consistency across measurements over time. The retrieved images were resampled to a uniform spatial resolution of 10m per pixel. Although Sentinel-2 originally has 13 spectral bands, the GEE imagery includes auxiliary bands for quality assurance and cloud masking, totaling 26 bands. The data were filtered to contain less than 5\% cloud cover between January 1, 2017, and November 1, 2024 \cite{Eva2025thesis}. Sentinel-2 has a revisit time of approximately 5 days, with consistent overpasses above the Biesbosch between 10:00 and 11:00 local time, ensuring uniform illumination and minimal shadow interference.

\begin{figure}[htb]
    \centering
    \begin{subfigure}[b]{0.38\textwidth} 
        \centering
        \includegraphics[width=\textwidth]{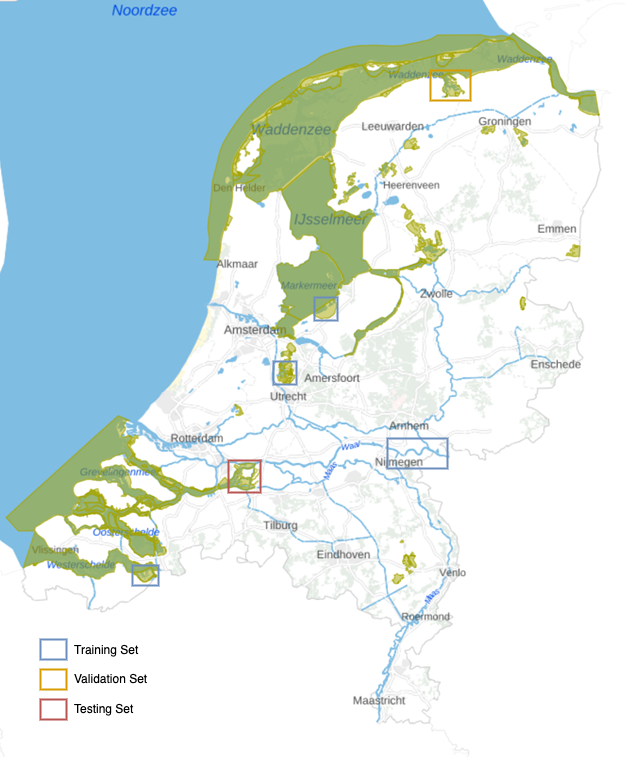}
        \caption{Train, validation and test set areas}
        \label{fig:wetland}
    \end{subfigure}
    \begin{subfigure}[b]{0.52\textwidth} 
        \centering
        \includegraphics[width=\textwidth]{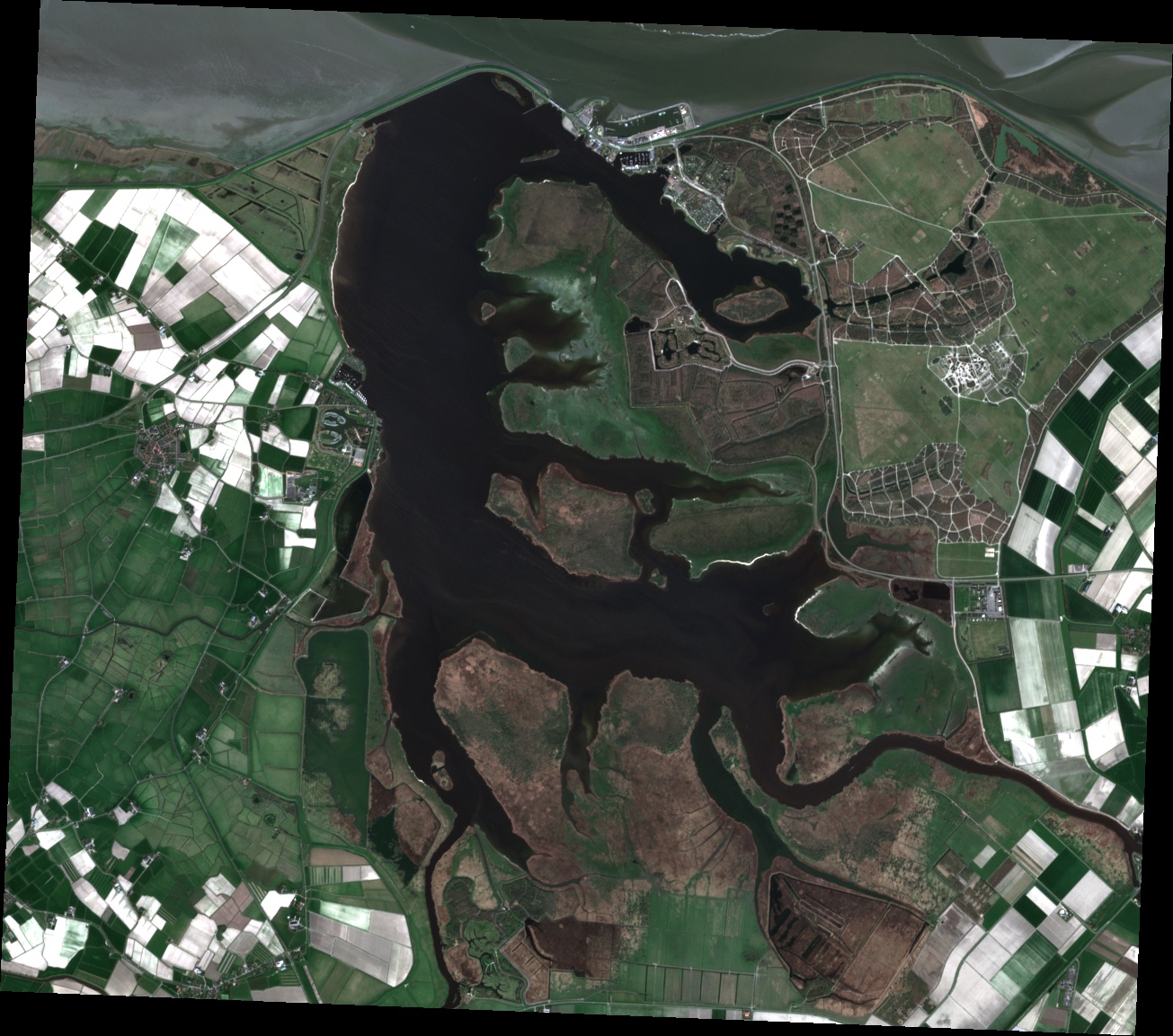}
        \caption{Image from the validation-set: Lauwersmeer}
        \label{fig:Lauwersmeer}
    \end{subfigure}
    \caption{\textit{\small Wetland areas included in the Sentinel-2 dataset}}
\end{figure}

The imagery of the Biesbosch area was used for testing, which meant that five similar wetland areas had to be selected for training and validation (see Fig.~\ref{fig:wetland}). As validation set the imagery from Lauwersmeer area (see Fig.\ref{fig:Lauwersmeer}) has been chosen, a Natura 2000 region in the north of the Netherlands. 

\paragraph{Medium-resolution land cover labels} for each Sentinel-2 image were sourced from the Dynamic World dataset, which provides near real-time, 10m resolution global land cover classifications using DL \cite{Brown2022DynamicMapping}. The dataset uses Sentinel-2 spectral data and a pretrained convolutional neural network trained on a combination of expert and non-expert annotations to assign probabilities for 9 land cover types to each pixel. The Dynamic World classification achieves an overall agreement of 73.8\% with expert-labeled validation data, performing best on categories like water, trees, and built areas while having more difficulty with classes such as grass and shrub \& scrub. For each pixel, the class with the highest probability (ranging from pixel values 0 to 8) is selected to create classification masks.


\vspace*{0.25cm}
\noindent
\textbf{Pléiades NEO imagery}\footnote{\scriptsize \url{https://earth.esa.int/eogateway/missions/pleiades-neo}}   was accessed via FTP from the Satellietdataportaal platform. This open-access platform provides high-resolution optical satellite imagery of the Netherlands and the Caribbean Netherlands and is only accessible to users with a Dutch IP address. It is offering spatial resolutions of up to 0.3m, significantly higher than Sentinel-2. Pléiades NEO provides 6 spectral bands: Red, Green, Blue (RGB), Near Infrared (NIR), Red Edge, and Deep Blue and has a revisit time of approximately 6 weeks.

Due to the high storage requirements of high-resolution imagery and the primary goal of the experiment to compare high and low-resolution data rather than evaluate the U-Net's performance, the focus was limited to the Biesbosch area. From this area, all available data was retrieved, spanning between the beginning of 2023 to the end of 2024. Furthermore, to ensure sufficient data availability while maintaining quality, a cloud cover limit of 30\% was applied during pre-processing. This threshold balanced data availability and minimized obstruction from clouds, ensuring clear visibility of the target features.

\paragraph{High-resolution land cover labels} for this dataset were manually created using the Beeldenboek bij Rijkswaterstaat\footnote{\scriptsize \url{https://open.rijkswaterstaat.nl/overige-publicaties/2020/beeldenboek-vegetatiebeheer-grote/}} as a reference, with the annotation process carried out in Roboflow\footnote{\scriptsize \url{https://roboflow.com}}.

\begin{figure}[ht]
    \centering
    \includegraphics[width=1\linewidth]{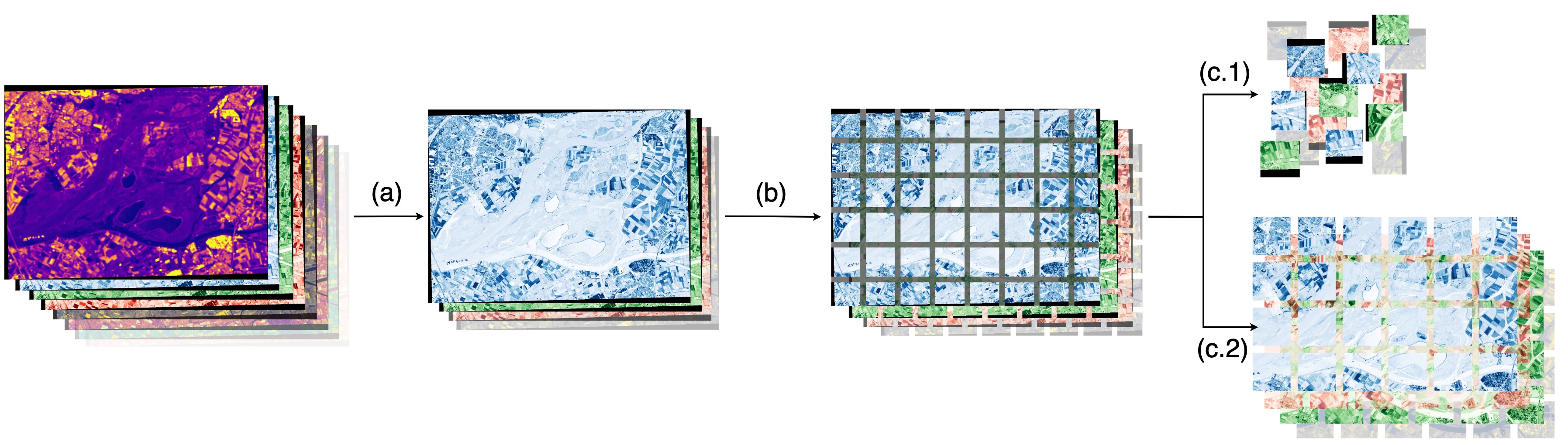}
    \caption{\textit{\small Overview of the pre-processing steps applied to both Sentinel-2 and Pléiades NEO datasets. (a) Selection of relevant spectral bands tailored to the classification task. (b) Division of images into patches of 256 $\times$ 256 pixels for medium resolution, or 1024 $\times$ 1024 pixels for very high resolution, for consistency and usability in training. (c.1) Exclusion of patches with dimensions smaller than the set size or containing excessive black pixels ($>10\%$ for Sentinel-2 and $>30\%$ for Pléiades NEO). (c.2) Retention of patches that meet size and quality requirements to ensure consistency in high-resolution data.}}
    \label{figure: data pre-processing}
\end{figure}

\begin{figure}[htb]
    \centering
    \includegraphics[width=1\linewidth]{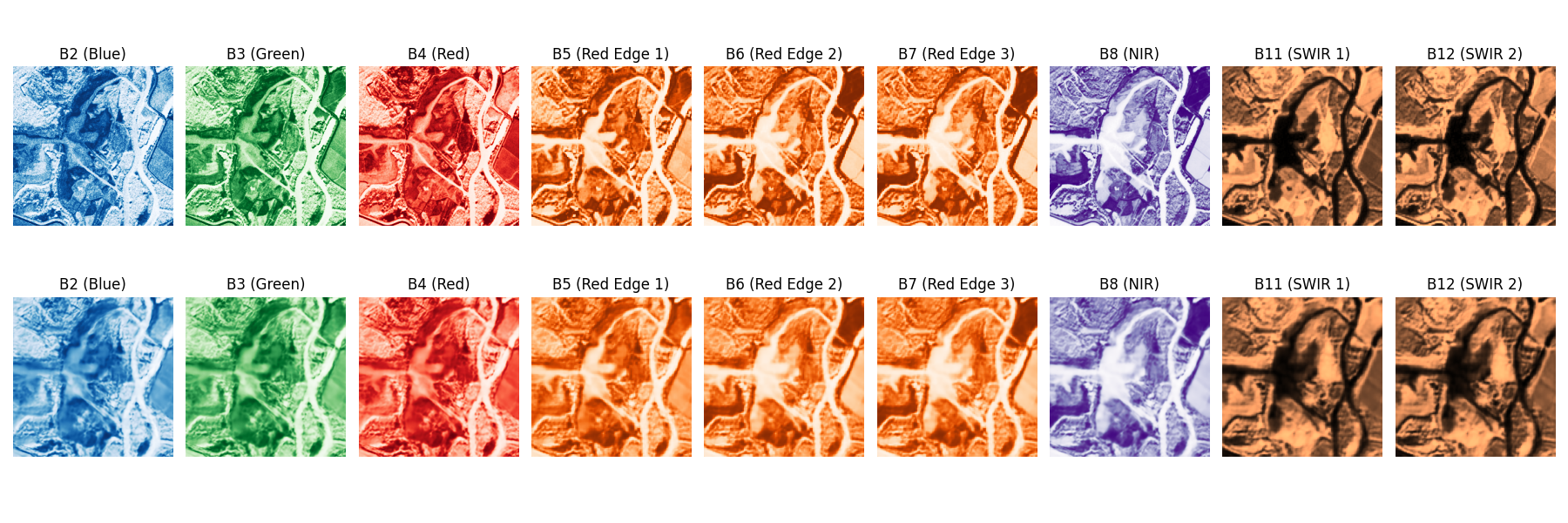}
    \caption{\textit{\small The reconstruction of the 9 individual spectral bands from Sentinel-2 imagery using an autoencoder. The top row is the original image, the bottom row the reconstructions The colors assigned to the bands are for intuitive representation, as the direct visual appearance of the bands differs from natural colors.}}
    \label{fig:9bandsen2}
\end{figure}

\vspace*{0.25cm}
\textbf{Data Pre-processing}
Before the data could be fed to the models, it underwent several pre-processing steps, as shown in Figure~\ref{figure: data pre-processing}. The same pre-processing pipeline was applied to both medium-resolution Sentinel-2 data and very-high-resolution Pléiades Neo imagery, ensuring diverse and representative data splits and enhancing the model's ability to generalize across different wetland environments. 
From the 26 available bands in the GEE Sentinel-2 imagery, 9 bands, as shown  in  Figure~\ref{fig:9bandsen2}, were selected based on their relevance to wetland classification tasks and the need to balance data size for computational efficiency. Comprehensive meta-analysis \cite{Jafarzadeh2022RemoteResearch,Mahdavi2018RemoteReview} on remote sensing for wetland classification shows that in addition to the RGB bands and SWIR-bands, the red-edge and near-infrared bands are the most effective optical bands for wetland delineation. Hence, the selected bands for the Sentinel-2 dataset are B2 (blue), B3 (green), B4 (red), B5-B7 (Red Edge 1-3), B8 (NIR), and B11-B12 (SWIR 1 and 2). These bands are known for their sensitivity to vegetation, water bodies, and soil characteristics, which are essential for distinguishing wetland environments. 

Due to storage limitations, only the first four spectral bands of the six available bands of the high-resolution Pléiades Neo dataset were used: Red, Green, Blue, and Near Infrared. When a direct comparison of the high-resolution Pléiades Neo the medium-resolution Sentinel-2 imagery was made, this was done for the same 4 spectral bands.

The Pléiades Neo data was tiled into \(1024 \times 1024\) pixel patches to account for its higher resolution and to capture more spatial detail within each patch, resulting in input dimensions of \(1024 \times 1024 \times n\), where \(n\) is the number of selected bands. Patches containing more than 30\% invalid pixels or those not meeting the required size were excluded from the dataset. The data was randomly split into 1,027 training images, 205 validation images, and 136 test images.


For a fair comparison, the medium-resolution images were divided into patches of size \(256 \times 256 \times n\). Here, patches containing more than 10\% black pixels or those not meeting the required size were excluded from the dataset. Lastly, the data was divided into three sets. The Biesbosch region was used as the test set to evaluate the model's performance on unseen data. Due to its similarity to the Biesbosch, Lauwersmeer was used as the validation set to tune model parameters. The remaining regions - Gelderse Poort, Oostvaardersplassen, Loosdrechtse Plassen, and Land van Saeftinghe - were used as the training set. Because the splits were based on geographical regions, which  differ greatly in size, the train/validation/test split was not balanced. A larger number of patches came from the retrieved  Biesbosch imagery, making the test set more extensive. The complete training set contained 1,701 images, the validation set 948 images, and the test set 1,140 images.

Lastly, histogram equalization was applied to the autoencoder inputs after the aforementioned pre-processing steps to further enhance feature differentiation and mitigate spectral inconsistencies.

\section{Experiment results}

\textbf{Autoencoder Reconstruction} The first experiment evaluates the ability of the autoencoder to reconstruct its input images. Performance is assessed on two datasets: medium-resolution imagery from Sentinel-2 and high-resolution imagery from Pléiades Neo.

\begin{figure}[htb]
    \centering
    \begin{subfigure}[t]{\linewidth}
        \centering
        \includegraphics[width=0.65\textwidth]{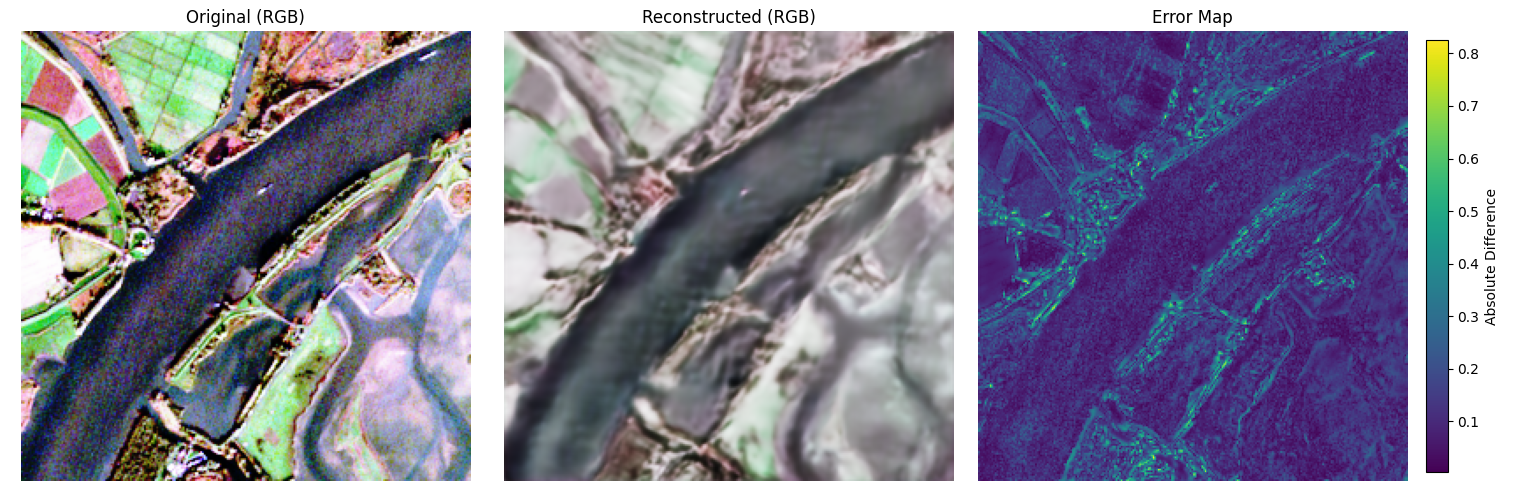}
        \caption{Reconstruction result for medium-resolution Sentinel-2 data.}
        \label{fig:sub1}
    \end{subfigure}
    
    \vspace{0.25cm}
    
    \begin{subfigure}[t]{\linewidth}
        \centering
        \includegraphics[width=0.65\textwidth]{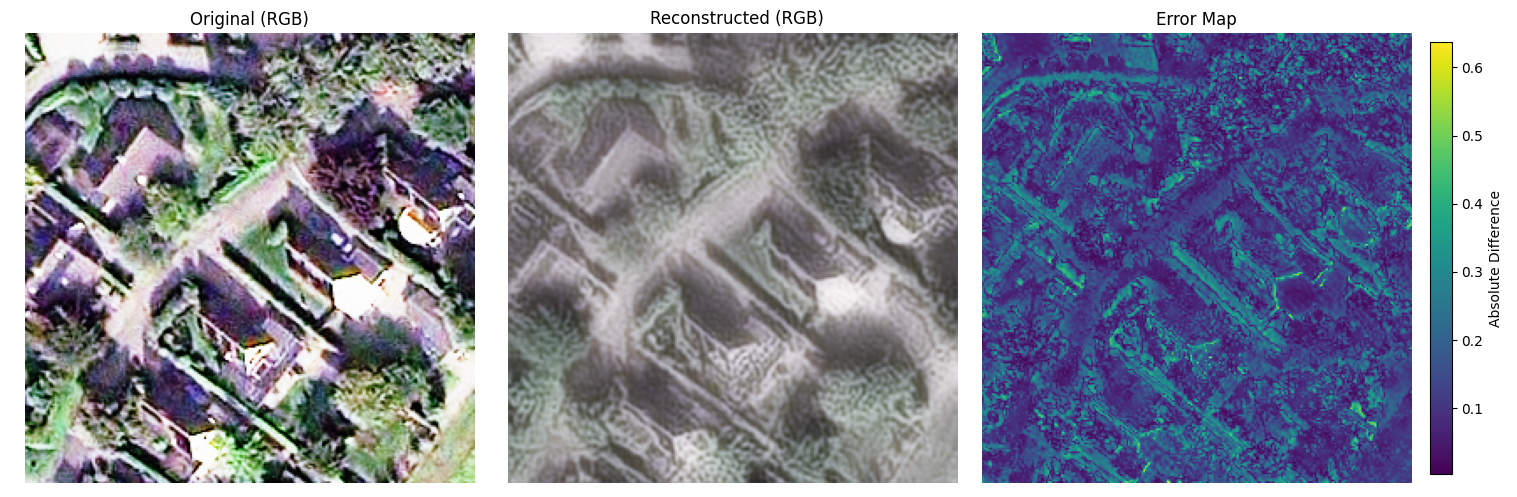}
        \caption{Reconstruction result for high-resolution Pléiades Neo.}
        \label{fig:sub2}
    \end{subfigure}
    
    \caption{\textit{\small Reconstruction results from the autoencoder for medium-resolution Sentinel-2 data (a) and  high-resolution  data (b). The first column shows the histogram-equalized original RGB image. The second column presents the reconstructed image from the autoencoder, while the last column displays the error map, where blue indicates minimal pixel differences and yellow highlights larger discrepancies.}}
    \label{figure: SDP and S2 reconstruction}
\end{figure}

The autoencoder is able to reconstruct the input images effectively, as illustrated in Figure~\ref{figure: SDP and S2 reconstruction} and Table~\ref{Table: autoencoderperformancee}. 
The training of the autoencoder for the medium-resolution imagery converged at 200 epochs, whereas for the high-resolution imagery, convergence was reached at just 9 epochs. 

\begin{table}[htb]
    \centering
    \begin{tabular}{lccccccc}
        \hline
        \textbf{Resolution} & \textbf{Accuracy $\uparrow$} & \textbf{PSNR $\uparrow$} & \textbf{SSIM $\uparrow$} & \makecell{\textbf{Huber $\downarrow$} \\ \textbf{Loss}} & \makecell{\textbf{SSIM $\downarrow$} \\ \textbf{Loss}} & \makecell{\textbf{Edge $\downarrow$} \\ \textbf{Loss}} & \makecell{\textbf{Mixed $\downarrow$} \\ \textbf{Loss}} \\
        \hline
        Sentinel-2 & \textbf{0.6076} & \textbf{17.8110} & \textbf{0.4964} & \textbf{0.0083} & \textbf{0.2518} & \textbf{0.1060} & \textbf{0.1009} \\
        Pléiades Neo & 0.3667 & 14.5375 & 0.4627 & 0.0111 & 0.2686 & 0.2342 & 0.1610 \\
        \hline
    \end{tabular}
    \vspace{0.25cm}
    \caption{\textit{\small Performance metrics of the autoencoder for medium-resolution (Sentinel-2) and high-resolution (Pléiades Neo) imagery on their corresponding test sets.}}
    \label{Table: autoencoderperformancee}
    \vspace{-20pt}
\end{table}

Table~\ref{Table: autoencoderperformancee} summarizes the performance of the autoencoder on the two datasets. Although the reconstruction results shown in Figure~\ref{figure: SDP and S2 reconstruction} are visually equally good, the numerical performance appears significantly lower for the Pléiades Neo.


\vspace{0.25cm}

\begin{figure}[hbt]
    \centering

    \begin{subfigure}[t]{\linewidth}
        \centering
        \includegraphics[width=\textwidth]{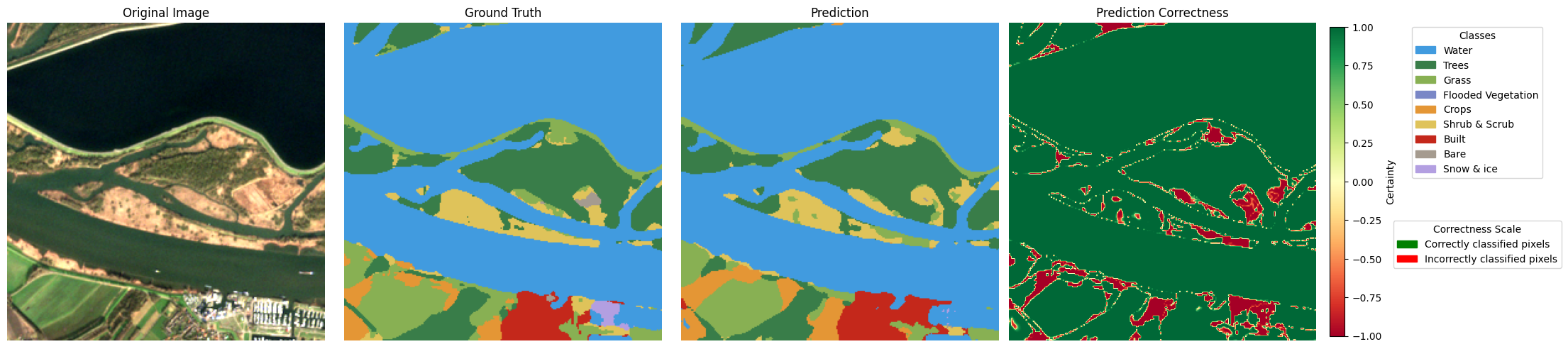}
        \caption{Land-cover classification for Sentinel-2 data trained from scratch. }
        \label{fig:sub2-1}
        \vspace{20pt}
    \end{subfigure}
    
    \begin{subfigure}[t]{\linewidth}
        \centering
        \includegraphics[width=\textwidth]{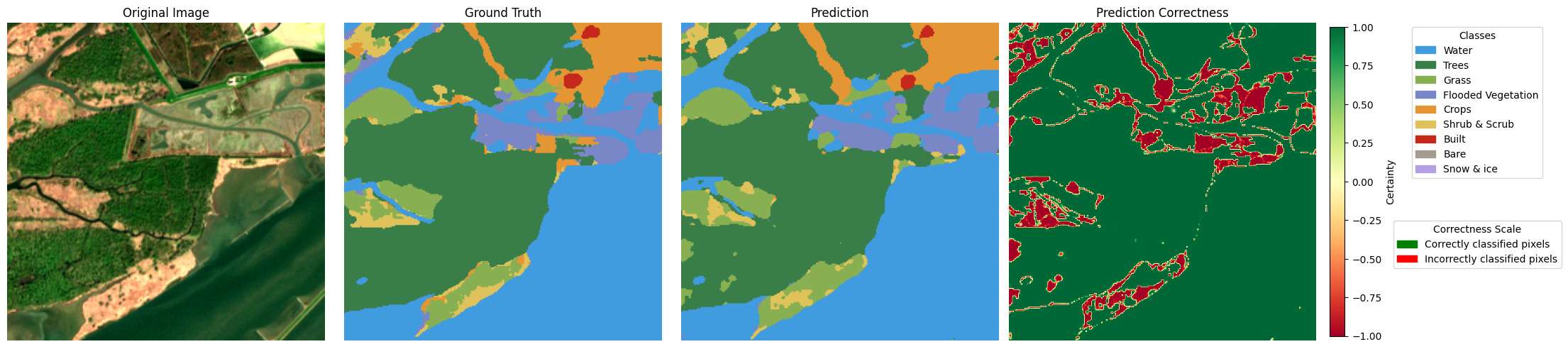}
        \caption{Land-cover classification for Sentinel-2 data with a pretrained model.}
        \label{fig:sub2-2}
    \end{subfigure}
    
    
    \caption{\textit{\small Land-use classification with U-Net for medium-resolution Sentinel-2 data with a model trained from scratch (a) and pretrained (b). The first column shows the histogram-equalized original RGB image. The second column presents the ground truth classification by Dynamic World.
    The third column shows the prediction of the U-Net. The last column displays the error map, where red represents misclassified pixels, and the color intensity reflects the model’s certainty.}}
    \label{figure: Unet segmentation maps}
\end{figure}

\textbf{Impact of pretraining on U-Net} The second experiment studies the effect of pre-training on the performance of the U-Net. Instead of using the pre-trained weights, the U-Net is allowed to train from scratch on the datasets from the four other Dutch wetlands, validated on the Lauwersmeer region, and tested on the Biesbosch region. The training from scratch converged after 300 epochs. These results are compared to a U-Net model trained for the same number of epochs, but initialized with the weights from the previously discussed autoencoder. The results are illustrated in Figure~\ref{figure: Unet segmentation maps}. Visually, the results are equally good again. However, the numerical performance between the two training methods is now also comparable (see Table~\ref{tab:unet_comparison100}). In this table, one can see a slightly higher recall for the model trained from scratch and a slightly higher precision for the pretrained model.

\vspace*{-1.5em}

\begin{table}[hbt]
\centering
\footnotesize
\begin{tabular}{@{}lccccccc@{}}
\toprule
\textbf{Training method} &  \textbf{Accuracy $\uparrow$} & \textbf{Dice $\uparrow$} & \textbf{IoU $\uparrow$} & 
\textbf{Precision $\uparrow$} & \textbf{Recall $\uparrow$} & \textbf{Dice Loss $\downarrow$} \\ 
\midrule
Scratch
     & 0.8526 & 0.6480 & 0.5346 
     & 0.6616 & \textbf{0.6694} & \textbf{0.4865} \\ 
Pretrained
     & \textbf{0.8542} & \textbf{0.6518} & \textbf{0.5378} 
     & \textbf{0.6923} & 0.6483 & 0.4905 \\ 
\bottomrule
\end{tabular}
\vspace{0.25cm}
\caption{\textit{\small Land-use classification performance for the U-Net model on the Sentinel-2 dataset with and without pretraining. 
The U-Net was trained for 300 epochs on the dataset. 
The pretrained model started with weights extracted from the autoencoder.}}
\label{tab:unet_comparison100}
 \vspace{-20pt}
\end{table}

The effect of pretraining was also evaluated on the high-resolution imagery from Pléiades Neo, using corresponding manually annotated labels (since the ground truth from Dynamic World is not available at this resolution). A clear improvement was observed, with, for instance, accuracy increasing from 60.35\% (scratch) to 88.23\% (pretrained), as shown in Table~\ref{Table: SDP comparison}.

\begin{table}[hbt]
\centering
\footnotesize
\begin{tabular}{@{}lccccccc@{}}
\toprule
\textbf{Training method} & \textbf{Accuracy $\uparrow$} & \textbf{Dice $\uparrow$} & \textbf{IoU $\uparrow$} & \textbf{Precision $\uparrow$} & \textbf{Recall $\uparrow$} & \textbf{Dice Loss $\downarrow$} \\ 
\midrule
Scratch
& 0.6035 & 0.2827 & 0.2243 & 0.3889 & 0.3158 & \textbf{0.5114} \\
\midrule
Pretrained
 & \textbf{0.8823} & \textbf{0.4457} & \textbf{0.3919} & \textbf{0.5079} & \textbf{0.4551} & 0.5457 \\  
\bottomrule
\end{tabular}
\vspace{0.25cm}
\caption{\textit{\small Performance comparison of non-pretrained and pretrained U-Net on high-resolution imagery and labels.}}
\label{Table: SDP comparison}
 \vspace{-20pt}
\end{table}

\vspace{0.25cm}

\begin{figure}[htb]
    \centering
    \includegraphics[width=0.95\linewidth]{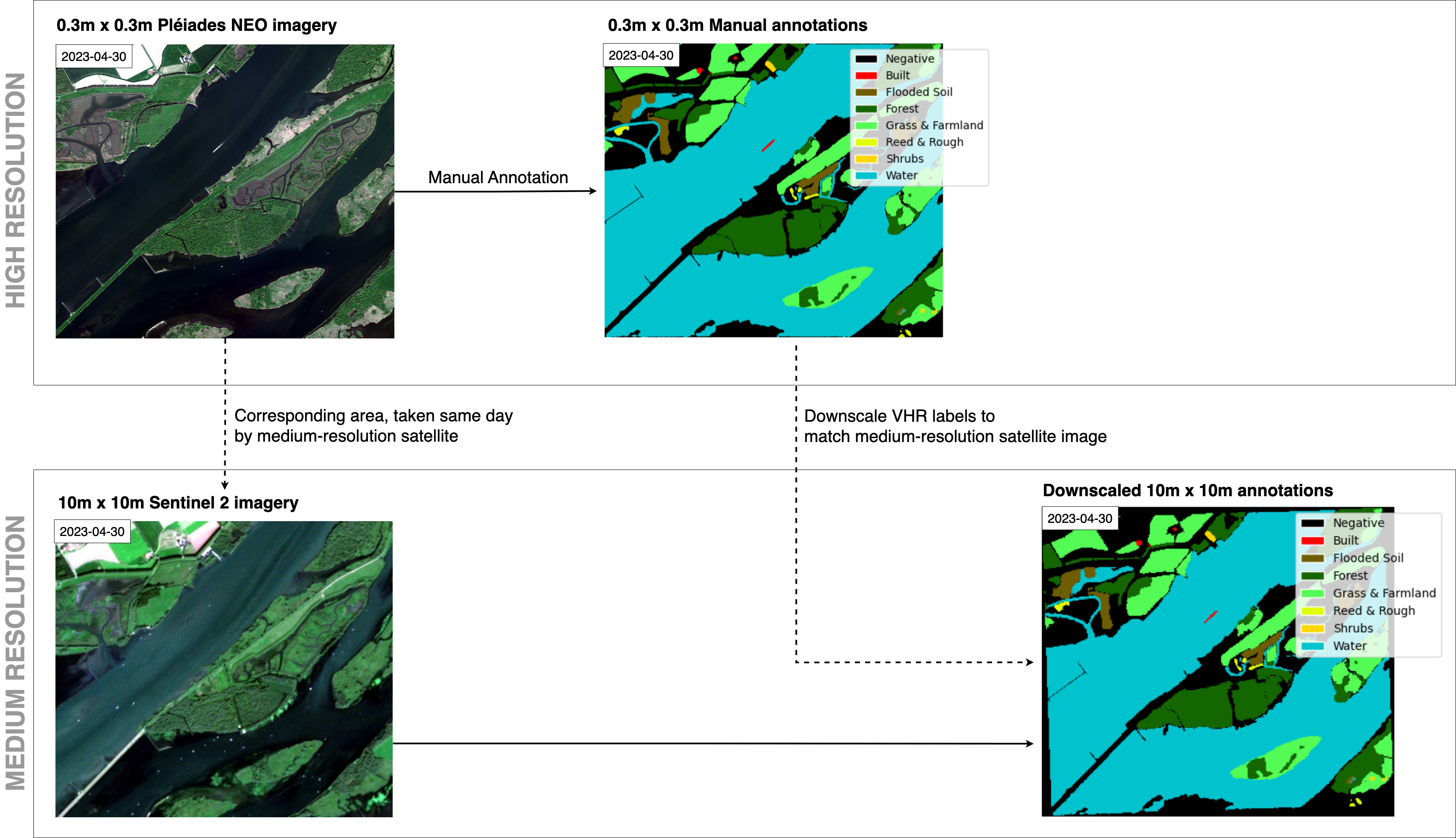}
    \caption{\textit{\small High-resolution imagery (VHR) from Pléiades NEO (0.3m x 0.3m) was manually annotated to generate high-resolution land cover labels. A corresponding medium-resolution Sentinel-2 image (10m x 10m) from the same date was found to ensure temporal consistency. The high-resolution annotations were then downscaled to match the lower resolution of the Sentinel-2 image. This process was repeated to generate multiple VHR and medium-resolution image-label pairs.}}
    \label{fig: labeltransfer}
\end{figure}

\textbf{Impact of Resolution} The third experiment studies the effect of the resolution on the land cover segmentation and classification of the U-Net. To achieve this, datasets from two sources were utilized: high-resolution images (0.3m $\times$ 0.3m) from Pléiades NEO and medium-resolution images (10m × 10m) from Sentinel-2. To ensure a fair comparison, the corresponding dates  used for the Pléiades NEO and Sentinel-2 images were selected to be temporally close. The high-resolution satellite images were divided into a 4 × 4 grid of patches, whereafter the 16 tiles were manually labeled using Roboflow. 
These manually annotated labels were transferred from the high-resolution imagery to the corresponding low-resolution imagery, and down-sampled, as shown in Figure~\ref{fig: labeltransfer}. By following these steps, the experiment aimed to assess the relationship between image resolution and the ability to classify detailed vegetation features, using consistent pre-processing of image-label pairs and training steps for both resolutions to ensure a controlled comparison. 

\begin{figure}[htb]
    \centering
    \subfloat[Medium-resolution segmentation results]{%
        \includegraphics[width=0.95\textwidth]{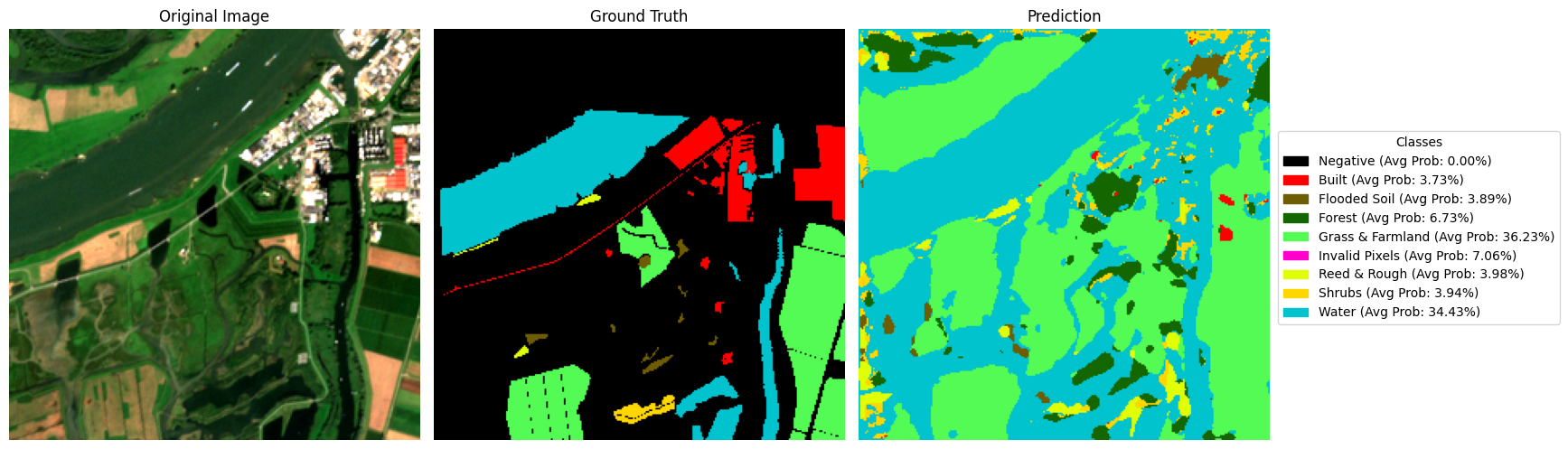}%
        \label{fig:subfig1}
    }\\
    \vspace*{0.25cm}
    \subfloat[High-resolution segmentation results]{%
        \includegraphics[width=0.95\textwidth]{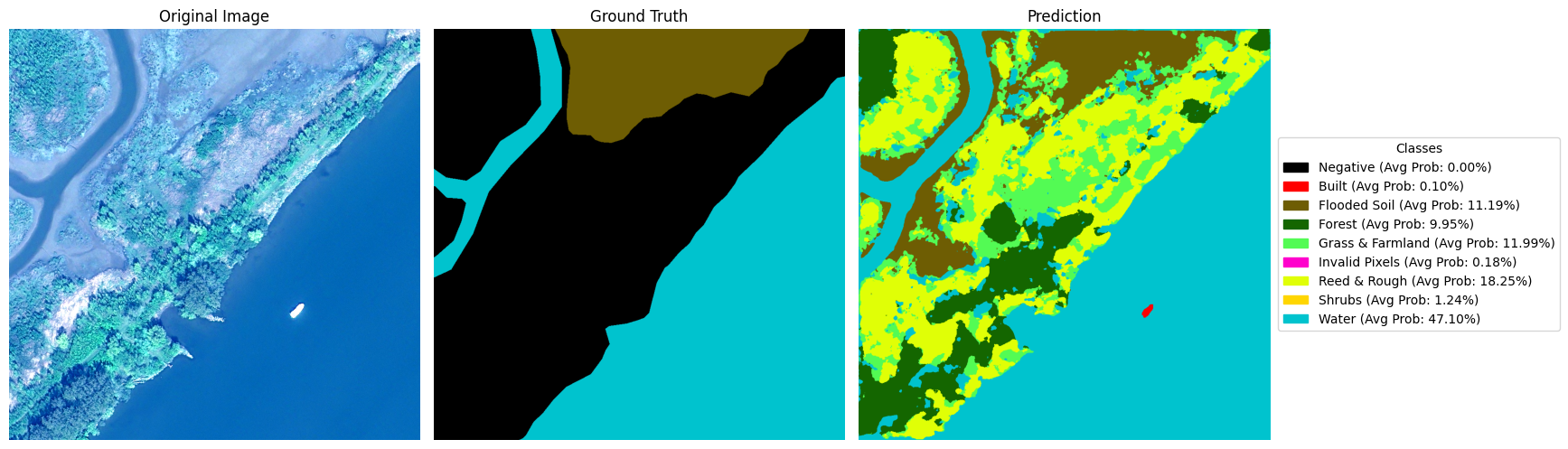}%
        \label{fig:boat}

    }
    \caption{\textit{\small Comparison of medium-resolution and high-resolution segmentation results. While performance metrics appear similar, high-resolution imagery (b) provides finer details and more precise segmentation. The legend shows the average predicted probability for each class across the image, highlighting model confidence in class assignments.
    \\}}
    \label{figure: vis experiment 2}
    
\end{figure}

Table~\ref{Table: resolution comparison} compares the performance of non-pretrained high-resolution labels with non-pretrained medium-resolution labels on both test sets. However, a more effective way to compare the performance of both approaches is through visual evaluation. Figure~\ref{figure: vis experiment 2} shows the results of semantic segmentation on both resolutions. 

\vspace{0.5cm}

\begin{table}[htb]
\centering
\footnotesize
\begin{tabular}{@{}lccccccc@{}}
\toprule
\textbf{Resolution} & \textbf{Weighted $\uparrow$} & \textbf{Dice $\uparrow$} & \textbf{IoU $\uparrow$} & \textbf{Precision $\uparrow$} & \textbf{Recall $\uparrow$} & \textbf{Dice Loss $\downarrow$} \\ 
\midrule
Medium 
& 0.5751 & \textbf{0.3454} & \textbf{0.2687} & \textbf{0.4148} & \textbf{0.4395} & 0.7241 \\
\midrule
{High} 
 & \textbf{0.6035} & 0.2827 & 0.2243 & 0.3889 & 0.3158 & \textbf{0.5114} \\  
\bottomrule
\end{tabular}
\vspace{0.25cm}
\caption{\textit{\small Performance comparison between medium-resolution (Sentinel-2) and high-resolution (Pléiades Neo) imagery.}}
\label{Table: resolution comparison}
\end{table}

From Figure~\ref{figure: vis experiment 2}, it is clear that, although the performance metrics for both resolutions appear comparable, and in some cases even slightly higher for the medium-resolution imagery and labels, the high-image imagery and labels provide more precise segmentation results. Note, for instance, in Figure~\ref{fig:boat} the small white boat in the water, which is classified as a built. Also Grass, Reed and Forest classes give reasonable predictions, although the ground truth concentrated on the larger areas that could be classified with a high confidence. This reflects the difficulty of getting reliable labels for high-resolution imagery, and the need for self-supervised methods.

\section{Conclusion}

The baseline U-Net demonstrates that land-cover classification is possible for medium-resolution imagery of Sentinel 2. The labels provided by the Dynamic World dataset are detailed enough to train the baseline U-Net to an accuracy of 85.26\%. 

For medium-resolution imagery the effect of pretraining the U-Net with the weights of an autoencoder is minimal.
The benefit of pre-training becomes more evident for high-resolution imagery of Pléiades Neo. For high-resolution imagery, where detailed labels are more difficult to obtain, pretraining can boost the accuracy from 60.35\% to 88.23\%.

Based on visual inspection, medium-resolution imagery allows for a clear distinction of categories such as water, trees, and built-up areas, but it remains difficult to separate classes like grass and shrub \& scrub. In high-resolution imagery, more detail appears to be available to distinguish between different vegetation types and small objects. While the land-cover segmentation and classification model produces reasonable predictions, the absence of high-quality, detailed labels makes it difficult to distinguish meaningful quantitative results from potential hallucinations.

Overall, it is clear that land-cover segmentation and classification of dynamic wetlands can be achieved through the combined use of self-supervised pretraining and high-resolution imagery, thereby reducing reliance on extensive annotated data. As an additional contribution, this work provides a curated and openly accessible Sentinel-2 dataset with Dynamic World labels, supporting reproducibility and future research in wetland classification.


\bibliographystyle{splncs04}
\bibliography{references, manual_references, paper_references}

\end{document}